\documentclass[journal]{IEEEtran}
\usepackage{subfigure}
\usepackage{amsmath}
\usepackage{amssymb}
\usepackage{amsfonts}
\usepackage{graphicx}
\usepackage{url}
\usepackage{ccaption}
\usepackage{booktabs} 

\ifCLASSINFOpdf
\else
\fi
\begin{document}
%
\title{Skeleton-based Approaches based on Machine Vision: A Survey}
%
%
%

\author{Jie~Li,~\IEEEmembership{Member,~IEEE,}
        Binglin~Li,
        and~Min~Gao
\thanks{J. Li and M. Gao are with Chongqing University of Science and Technology, Chongqing, China, 401331 e-mail: 2014008@cqust.edu.cn.}
\thanks{B. Lin is with Chongqing University, Chongqing, China, 400001}
}

%
%

\markboth{}%
{Shell \MakeLowercase{\textit{et al.}}: Bare Demo of IEEEtran.cls
for Journals}
%



\maketitle

\begin{abstract}
Recently, skeleton-based approaches have achieved rapid progress on the basis of great success in skeleton representation. Plenty of researches focus on solving specific problems according to skeleton features. Some skeleton-based approaches have been mentioned in several overviews on object detection as a non-essential part. Nevertheless, there has not been any thorough analysis of skeleton-based approaches attentively. Instead of describing these techniques in terms of theoretical constructs, we devote to summarizing skeleton-based approaches with regard to application fields and given tasks as comprehensively as possible. This paper is conducive to further understanding of skeleton-based application and dealing with particular issues.
\end{abstract}


%
\IEEEpeerreviewmaketitle

\section{Introduction}
\label{03.03-03.09}
Skeleton-based approaches, also known as kinematic techniques, cover a set of joints and a group of limbs based on physiological body structure. Typically, the number of joints is determined by computational complexity, which is between ten to thirty. In recent years, some significant techniques have worked successfully on discovering the representation of skeletons with dots and edges, which can be categoried as top-down methods (e.g., cascade pyramid network \cite{chen2018cascaded} and PoseFix \cite{moon2019posefix}) and bottom-up methods (e.g., Openpose \cite{cao2018openpose}).

In order to analyze skeleton-based approaches deeply, we observe these researches from two aspects(i.e., single-frame and multi-frame). Fig. 1 describes the summarization. In regard to single-frame type, tasks are handled through investigating independent images within videos. Correspondingly, multi-frame type requires a series of sequential images among videos to explore the inherence.

\begin{figure*}[htp]
	\centering
	\includegraphics[width=0.9\textwidth, height=0.4\textwidth]{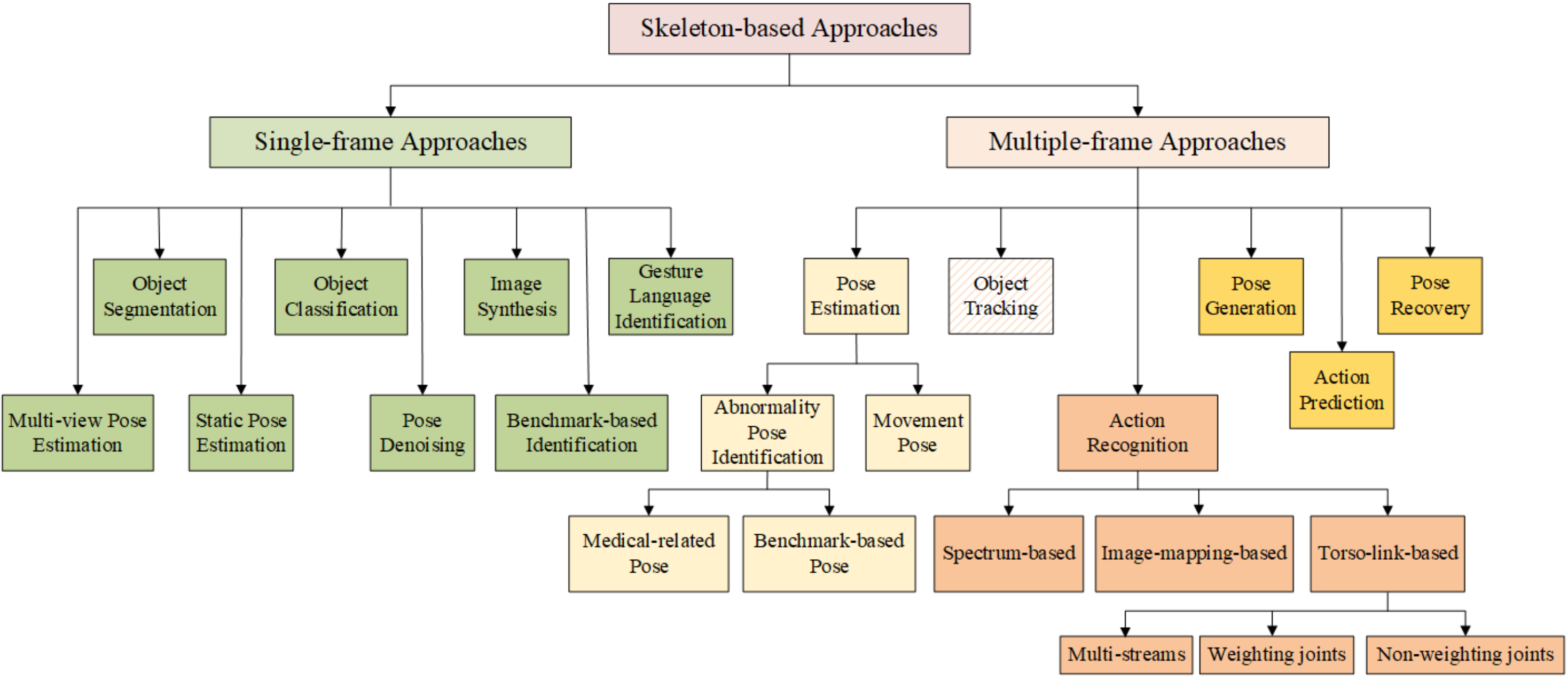}
	\caption{Types of skeleton-based approaches.}
\end{figure*}

\subsection{Single-frame Approaches}
\label{03.17-03.23}
\subsubsection{Multi-view Pose Estimation}
\label{}
Although skeletons with 3D structures enhance the accuracy of pose estimation, the training operation needs plenty of 3D ground-truth data which is costly. Thus, multiple 2D skeletons from different views are integrated into 3D structures under a given strategy to determine the pose. EpipolarPose \cite{kocabas2019self} implements epipolar geometry to combine the 2D skeletons and trains a 3D pose estimator with camera geometry information. RepNet \cite{wandt2019repnet} finds the mapping from 2D to 3D skeletons by designing an adversarial training approach and adding feedback projection from 3D to 2D skeletons. Based on Wasserstein generative adversarial network (WGAN), RepNet generates 3D skeletons by sending 2D skeletons as input into WGAN. The produced 3D skeletons are reprojected to 2D domain by camera loss. A substitutive fusion approach \cite{Nur2015AzizaAzis} tracks the status of joints in two views of 2D skeletons and labels the joints as "well tracked", "inferred", or "not tracked". For a joint in different views, the one diagnosed as "well tracked" has the highest priority, and then the inferred one is better than the one of "not tracked". Furthermore, dynamic time warping with K-nearest neighbor is adopted to classify learned skeleton representation. In regard to multiple persons, a multi-way matching algorithm \cite{dong2019fast} clusters detected 2D skeletons by sticking the keypoints of a person in various views. 

Two networks (named as VA-RNN and VA-CNN) work together to discover skeleton representation of actions \cite{Pengfei2017Zhang, Pengfei2019Zhang}. An automatic selection scheme is involved in both the nets to choose prior viewpoints of 2D skeletons rather than a fixed criterion. In VA-RNN, 2D skeletons are rotated to obtain a complex feature by training multiple LSTM. For VA-CNN, a intrinsic feature is abstracted by a convolutional network. These two features are fused to determine action classes.

Additionally, loss functions (e.g., regression loss and consistency loss.) in deep neural networks (e.g., CNN) are designed to improve pose estimation within multi-view skeletons, and distance biases of skeleton information in multiple views are minimized to improve fusion accuracy \cite{rhodin2018learning}. 

\textbf{Challenges:} Relation evaluation (e.g., similarity) between skeletons in two views is still an important issue.

\subsubsection{Object Segmentation}
\label{}
Object segmentation based on skeletons aims at cutting targeted items (including biological and non-biological things) from others. In this type of tasks, object positions should be ascertained first and then object margins are drawn out.

Pose2Seg \cite{zhang2019pose2seg} focuses on human segmentation in images. In the model, detected skeletons are compared with established standard skeletons of each pose, and a affine transformation matrix calculates similarity. SegModule is presented to understand skeleton features and corresponding visualization. In addition, semantics is introduced as a supplementary perspective. By training a Part FCNto obtain semantic part score map, the body is further divided into head, torso, left arm, left leg, right arm, and right leg \cite{xia2017joint}. Another human segmentation approach \cite{junior2012skeleton} discovers edge information through adding all widths of physical parts (e.g., head and shoulders, neck, chest, hip) into body skeletons. In consideration of cascade connection of curve skeletons like a tree and its leaves, body segmentation are realized by trimming branches and isolating intersectional leaves \cite{lovato2009automatic}. In a human counting case, blurry margins are allowed, let alone body parts. After removing background, all heads are identified within skeleton graph segmentation and utilized to gain human amount \cite{merad2010fast}. 

Besides of human body segmentation, animal and non-biological items are also being studied to be segmented. Critical points are selected on the basis of surface skeletons and expand to component sets separately \cite{reniers2007skeleton}. With skeleton matching techniques, skeletal branches over a particular position are reconstructed by estimating the distances between two views and used to reconstruct the non-biological things \cite{durix2016skeleton}.

\textbf{Challenges:} Pinpoint object margins of diverse items are confirmed not only depending on skeletons but also item structures.

\subsubsection{Static Pose Estimation}
\label{}
Without comparison with any benchmark, it is hard to evaluate poses by only one image.  

\textbf{(1) Single person}. 

A path (i.e., 2D-3D-2D) learns the features of skeletons severely through projecting into 3D and being reprojected into 2D with lifting networks. The original 2D skeleton is as input \cite{chen2019unsupervised}.  A parsing induced learner exploits parsing information to enhance skeleton information through a pose encoder. A pose encoder abstracts pose features while another pose encoder fuses residual information into pose representation \cite{nie2018human}. HRNet has parallel high-to-low resolution subnetworks to gain both high and low resolutions of skeletons and sums them up \cite{sun2019deep}. Based on ConvNets, 2D images are translated into 3D skeleton models. For instance, heatmaps and silhouettes are extracted from a 2D body, and then pose and shape parameters are abstracted separately. These parameters are meshed together into a 3D body with 2D annotations \cite{pavlakos2018learning}. In another case, depth knowledge is integrated with ConvNets between paired joints in skeletons to mix into a 3D pose \cite{pavlakos2018ordinal}. PoseRefiner implements skeletons in binary channels and pose classes to train ConvNets learning likelihood heatmaps to refine the skeletons \cite{fieraru2018learning}. A dual-source approach learns the representations both from 2D and 3D skeletons \cite{yasin2016dual}. 3D poses are projected into multiple 2D skeletons, and used to find the highest likelihood along with a test image. To address skeletons seriously, a upper-body visualization uses different colors and polylines to distinguish between the left and right body \cite{johnson2011learning}. With these representations, 16 poses are clustered with high accuracy. Furthermore, ConvNets parameters are analyzed to find prior settings to identify poses with skeletons \cite{rafi2016efficient}.

\textbf{(2) Multiple persons}. 

Cascaded Pyramid Network (CPN) containing two networks (i.e., GlobalNet and RefineNet) \cite{chen2018cascaded} adopts a top-down pipeline which means locating skeleton keypoints first with ResNet backbone, extracting features of these keypoints as HyperNet, and then assembling them. A multi-person pose estimation (RMPE) involves spatial transformer networks to rectify various ground truth bounding boxes for people and a final box is obtained for each person \cite{fang2017rmpe}. DeepCut \cite{pishchulin2016deepcut} uses adapted fast R-CNN (AFR-CNN) to detect body parts with integer linear programming and dense CNN (Dense-CNN) to obtain the intensity with geometric and appearant constraints. Additionally, based on DeepCut, DeeperCut \cite{insafutdinov2016deepercut} is proposed to improve body part detection with a bottom-up pipeline and an image-conditioned pairwise assembling strategy is designed. Angles among body parts are observed meticulously to assist in searching pairwise joints.

\textbf{Challenges:} Without consecutive images of an action, features learned from static poses can resolve the estimation.

\subsubsection{Object Classification}
\label{}
The purpose of object classification is to identify the items in images through observing skeletons without any limitations of object types. 

In order to offset skeleton noise in classification, skeleton contours are trimmed as a trade-off between shape reconstruction error and skeleton simplicity based on Bayesian theory \cite{shen2013skeleton}. Tree representation of skeletons as a graph is turned into strings of skeleton edges, and a deformable contour method is used to compare these strings of diverse objects \cite{he2004graph}. Node pairs in curve skeletons are obtained by cascade of symmetry filters and a symmetry correspondence matrix is designed to gain symmetry cloud which is classified by spectral analysis \cite{jiang2013skeleton}.

\textbf{Challenges:} Curve skeletons are main tools for object classification with obvious transformation of each object.

\subsubsection{Pose Denoising}
\label{}
Typically, skeletons are dominant for special structures, especially for human body. When odd poses occur and parts of bodies are overlapped, it is hard to peel skeletons of each body. Redundant joints and limbs are deemed as noise, which can be eliminated by filtering linear transformations \cite{Girum2018Demisse} and comparing joint positions and limb angles with standard ones \cite{LiChi2019Liao}.

\textbf{Challenges:} Advanced denoising algorithms may be used to further improve the capability of pose denoising.

\subsubsection{Image Synthesis}
\label{}
Commonly, image synthesis tries to produce other views of skeletons by learning the representation of skeletons in a view. GAN is a popular technique to solve this problem, but traditional GAN has weak ability of obtaining the relationship among joints and limbs. Bidirection GAN is adopted to search the mapping from an initial pose skeleton to another pose skeleton  \cite{pumarola2018unsupervised}. Deformable GANs \cite{siarohin2018deformable} use heatmaps from skeletons as conditional information and human poses as originial images feeding into the generator to generate other pose images. The generated poses are shifted into the discriminator with corresponding heatmaps. Furthermore, CRF-RNN  is established to predict conditional human body with pose transforamtion from a given skeleton to a target skeleton \cite{si2018multistage}. Mask R-CNN model is built to transform a pose skeleton into another pose skeleton, and paralleling models operate simultaneously with diverse keypoints sampled from initial pose skeletons. These mappings are combined together to creat a new pose skeleton \cite{radosavovic2018data}. Some researches conduct physical structure characteristics to reconstruct images other than studying the inherence. For example, body skeletons are divided into multiple components and the background is drawed out \cite{balakrishnan2018synthesizing}. These components are rotated and integrated into a new body which is blended into a synthesized new background.

\textbf{Challenges:} GAN is a useful tool in generation although its performance is still unstable. While applying GAN on skeleton synthesis, how to improve the quality is still a challenge.

\subsubsection{Benchmark-based Identification}
\label{}
Comparison with benchmark is a straight way to judge the class of items. For skeletons, key points in standard and objective skeletons are contrasted in turn. 

Skeleton information (joints and limbs) is transformed into a tree with key points which are compared with given trees. The best match of two shape trees are treated as identical thing \cite{Bo2009Jiang}. A skeleton graph of any object is also used to contrast with standard units in both high geometric and topological similarities \cite{sundar2003skeleton}. To highlight the joints in comparisons, lines among joints address the relations of two joints instead of limbs. Unlike a fixed number of compared joints in previous two techniques, comparison steps in skeleton graphs here are random \cite{giang2013skeleton}. Similarity metrics along with 3D skeleton features are designed to obtain the similarity between the human skeleton in a single frame image and the templates \cite{munaro2014feature}. Body dot clouds are cooperated with main curve skeletons to judge a matching ratio of human motions, which works better than the case only involved curve skeletons \cite{hajdu2007object}. Skeletons using Kinect from upper and lower body are analyzed separately to identify human gaits with ANN as a classifier \cite{sinha2013person}.

\textbf{Challenges:} The evaluation criterion of similarity between the object and template skeletons is controlled by experience. Complex metrics may play well on similarity assessment.

\subsubsection{Gesture Language Identification}
\label{}
Body gestures have rich information as well as language, also called as body language. Deep learning algorithm is a novel tool to understand gesture meanings, such as RNN \cite{Zhi2017Zhang} and LSTM \cite{Xin2019Liu}. Deep networks extract features of gesture skeletons, which is easier than those of intact bodies.

\textbf{Challenges:} Unlike full body skeletons, most body gestures only contain partial joints and limbs. Thus, application scans should be settled to ensure the involved groups of joints and limbs.

\subsection{Multi-frame Approaches}
\label{03.23-03.29}
\subsubsection{Dynamic Pose Estimation}
\label{}

\textbf{(1) Abnormality pose identification.} 

\textbf{1) Medical-related pose.} A view adaptive LSTM (VA-LSTM) aims at detecting medical condition (i.e., sneeze/cough, headache, neck pain, staggering, chest pain, vomiting, falling, and back pain), containing a classification and regression subnetwork \cite{yin2019skeleton}. Original skeletons are rotated and translated into new architectures and then sent into the subnetworks to learn the corresponding medical classes. Neuromusculoskeletal disorders are also detected by pose estimation \cite{elkholy2019efficient}. Asymmetry features are extracted with splitted results of body joints according to the left and right body, which is used to catch normal motion patterns by a probabilistic normalcy model. The likelihood between a test action and a normal motion is computed to determine the abnormality.

\textbf{2) Benchmark-based pose.} Deep learning structures (e.g. CNN \cite{wu2019skeleton} and LSTM \cite{jeong2019human}) are applied to abstract the features which are compared with a series of skeleton benchmarks (i.e., joints and limbs).

\textbf{(2) Movement pose.} 

Graph convolutional network (GCN) is a core solution for movement pose estimation on account of its strong ability of capturing spatio and temporal features \cite{zhao2019semantic, zhang2019graph, gao20193d, ryu2019skeleton, li2019novel, gao20193d}. Except GCN, deep learning structures (e.g., CNN \cite{pavllo20193d, dhiman2019skeleton}, RNN \cite{chen2017motion, pavllo20193d}, and LSTM \cite{feng2019spatio}) are also key techniques for movement pose estimation through learning representations under given conditions.

Moreover, physical analysis of skeletons and probability estimation are also utilized in movement pose estimation. A exemplar-based method is explored to adjust initial estimated poses with inhomogeneous systematic bias while skeletons are defined as a simple directed graph and limbs are directed arrows. A regression function is proposed to predict the pose with rooted-mean-squared differences between templates and objective skeletons \cite{shen2012exemplar}. 2D keypoints extracted by pose estimators with skeletons are fused with SMPL regressors to create 3D models with accurate camera parameters \cite{arnab2019exploiting}. Semantic representations of volume occupancy and ground plane support are helpful for distinguishing multiple persons after evaluating each single person with 3D skeletons \cite{zanfir2018monocular}. On the strength of spatial positions and joint changes, a decision tree can quickly recognize basic action events and monitor action types under graph constraints of state transition \cite{han2013real}.

\textbf{Challenges:} Unlike static pose estimation, dynamic pose estimation usually discusses both spatio and temporal features with a series of images. The mixture process is a key point to control the quality of final representations.

\subsubsection{Object Tracking}
\label{}

PoseTrack \cite{andriluka2018posetrack, iqbal2017posetrack} follows the particular persons in videos, including multi-person pose estimation in a image, multi-person pose estimation in videos, and multi-person articulated tracking. ArtTrack \cite{insafutdinov2017arttrack} draws body part graphs in temporal aspect and abandons joints with loose relation by a feed-forward convolutional architecture. For robot, visual distances and lighting intensity are considered while following human using Kinect  \cite{shukor2016scene}. Also based on Kinect and Kalman filter, foot points are detected and followed with depth information and pairwise curve matching in 3D space from given views to a virtual bird’s eye view \cite{kuo2013skeleton}. With capturing keypoints in motion and fitting skeleton, human pose is tracked with transformed 3D models \cite{herda2001using}. Unlike deep learning structures, images in 3D models are filtered and the likelihood is calculated between shape models deformed by skeleton poses and image data with regard to probability theory \cite{shaheen2009comparison}. Fast movements of animals and persons are traced with non-rigid temporal deformation of 3D surface \cite{gall2009motion}. Other than tracking human, people handling objects are traced with GCNs after detecting hand joints in body skeletons \cite{kim2019skeleton}. 

\textbf{Challenges:} The relation of a given object between sequences is a crucial issue in object tracking.

\subsubsection{Action Recognition}
\label{}
Action recognition is a crucial part of skeleton approaches, which has been successfully applied to a lot of real projects. Among action recognition techniques about skeletons, there are three categories: (1)Spectrum-based, (2)Image-mapping-based, and (3)Torso-link-based approaches. In torso-link-based techniques, three subtypes are involved: 1) Multiple Streams, 2) Weighting joints, and 3) Non-weighting joints.

\textbf{(1) Spectrum-based}.

Skeleton sequences are turned into color spectrum dots with ConvNets, involving a procedure with 1) joint distribution mapping, 2) spectrum coding of joint trajectories and body parts, and 3) joint velocity weighted saturation and brightness. The spectrum actions are transferred from different filmed angles and then sent to the corresponding ConvNets which separately give scores with regard to actions \cite{hou2016skeleton}. 

\textbf{(2) Image-mapping-based}.

In this area, action skeletions are tranformed into feature maps which are learned by convolutional neural networks as the main solution backbone. With a single CNN, both temporal and spatial information obtained by physical information of joints and limbs are transformed into representation images which are sent to a CNN-family model (e.g., VGG \cite{li2017end} and CNN-LSTM \cite{hoang20193d}) for learning features \cite{du2015skeleton, pham2018learning}. Furthermore, the representation images are stretched with various sizes and fed into multiple CNNs to combine a highly representative explanation, in which these CNNs have the same structure \cite{li2017skeleton} or diverse structures \cite{cao2018skeleton}. Feature maps in traditional three coordinate dimensions (i.e., X, Y, and Z axis) of 3D skeletons are dissembled to corresponding CNNs, resulting in multi-column representations of actions \cite{nguyen2017multi}. Besides of coordinate dimensions and time dimension in 3D action skeletons, color (i.e., RGB) is also deemed as a dimension to learn deep features in action recognition \cite{liu20173d, laraba20173d}. Features of different factors of action skeletons (i.e., joint-joint distances, joint-joint orientations, joint-joint vectors, joint-line distances, line-line angles) obtained by physical computations are encoded into images and loaded into multiple CNNs to further extract features \cite{ding2017investigation, rostami2017skeleton, ren2018investigation}. For a joint in skeleton sequences, potential relations are explored by position chains of a physical body and mixed with features learned from each frame by CNN for compact representations of actions \cite{ke2017new}. Additionally, temporal consistency is deeply discovered by establishing  more networks to analyze extra information in actions \cite{luvizon20182d}. Thereinto, DD-net \cite{yang2019make} is a popular framework.

Unlike learning abstract features in action images, traditionally physical and mathematical techniques are still useful in action recognition under certain conditions. Physical information of joint changes is also used to detect actions through calculating precise relations (e.g., distances and angles) among joints for each action \cite{jiang2015informative, bloom2012g3d}. Geometrical relationship of limbs and joints draws a tree in which actions as a father node link key poses. The actions are viewed as tree nodes and features derived from those actions are as child nodes \cite{wang2014cross}.

\textbf{(3) Torso-link-based}. 

Torso-link (also called stick-link) is a most commonly used technique in skeleton-based approaches.

\textbf{1) Multiple Streams}.
More than two aspects of representations gained from torso-link skeletons are used to learn features of actions by dependent and parallel networks, e.g., spatial and temporal \cite{tu2018skeleton, shi2019two}, dots and lines \cite{zheng2019relational, li2019learning}, joints and time \cite{wang2017modeling, xu2018ensemble}, position and feature \cite{si2019attention}, spatial, temporal, structural, and actional \cite{wang2019robust}. The types of networks depend on action characteristics, e.g., RNN \cite{tu2018skeleton}, RRN \cite{tu2018skeleton, wang2017modeling}, CNN \cite{li2019learning, xu2018ensemble, wang2019robust}, GCN \cite{shi2019two}, and LSTM \cite{si2019attention}.

\textbf{2) Weighting joints}.
In other categories, all items in joints or/and limbs are given equal importance. However, in particular cases, actions have typical changes of partial joints and limbs which can represent the actions severely. Significant joints and limbs can be chosen by covariance matrix \cite{nguyen2018novel}, filtering function on the basis of skeleton graphs \cite{li2018action}, CNN \cite{hu2017body}, LSTM \cite{ding2015skeleton}, multi-head attention model with itereative attention on diverse parts of a body \cite{zhang2019multi}, projecting skeleton angles onto a unit sphere \cite{youssef2016spatiotemporal}, information gain with regard to position and velocity histogram from skeletons \cite{wang2018key}. Significance sorting techniques of all joints and limbs are conducive to reducing the hardness and complexity of skeleton feature extraction in action recognition, including spatial pyramid model (SPM) \cite{li2019novel}. Weighting the features of different parts of human body is also valid for identifying actions, e.g., bidirectional RNN \cite{du2015hierarchical, du2016representation}. 

\textbf{3) Non-weighting joints}. This part is the main component of action recognition, where overviews are sufficient over last decades \cite{poppe2010survey, weinland2011survey, holte2012human, chaquet2013survey, guo2014survey, zhu2016handcrafted, herath2017going, d2019survey, jobanputra2019human, chen2020monocular}. Therefore, we here introduce the sketch briefly. In the field, inherent representation of skeletons of actions is gained by two ways: physical computation and feature extraction. 

For the former, based on empirical knowledge of human body, the relations among joints and limbs under a particular action is calculated with explicit equations \cite{li2019skeleton, baptista2018key, ahmed2015kinect, taha2015skeleton, cippitelli2016evaluation, hbali2017skeleton, ding2015stfc}.  

For the latter, deep neural networks and other techniques devote to learning skeleton features for each action. Typically, deep neural networks have gained great success on action recognition, e.g., DNN \cite{weng2017spatio, huang2017deep, hu2019joint}, CNN \cite{nikolov2018skeleton, hosseini2019deep, zhu2019cuboid, ke2018learning}, RNN \cite{wang2018beyond, wei2017human, ren2018robust}, LSTM \cite{liu2017skeleton, jun2018deep, liu2017global, zhang2018fusing, cui2018multi, meng2019sample, tu2018spatial, liu2018learning, lee2017ensemble}, GCN \cite{yan2018spatial, ye2019joints, yang2020pgcn, liu2019si, gao2018generalized, shi2019graph, shi2019skeleton}. 

Moreover, traditional machine learning algorithms are designed to identify actions, e.g., kNN \cite{ubalde2016skeleton}, RBF \cite{yang2019spatio}. HMM discovers the semantic information of actions which assists in action identification \cite{yu2009human, ding2015skeleton}. Reinforcement learning is also a technique to obtain effective representation of an action \cite{zheng2018unsupervised}. Bayesian varies across different sequences of an action \cite{zhao2019bayesian}.

Additionally, probability and mathematical theories are useful in action recognition, e.g., analogical generalization and retrieval \cite{chen2018action}, screw matrices \cite{ding2017skeleton}, gradient vector flow comparison \cite{yoon2013human}, discriminative metric \cite{chaudhry2013bio}.

\textbf{Challenges:} In action recognition, the greatest challenge is to determine the start and end moment of an action. Usually, fixed time interval is adopted and leads to high deviation while actions have widely different time intervals. 

\subsubsection{Action Prediction}
\label{}

Unlike HMM, CRF, RNN, LSTM, CNN, a latent global network with latent long-term global information is designed to predict an action \cite{ke2019learning}. Based on the competition in GAN, two nets (i.e., I-Net and D-Net) are trained iteratively. Full and partial sequences are sent into I-Net to learn representations, separately. Afterwards, the representations are distinguished by D-Net.

\textbf{Challenges:} The evaluation of likelihood between the existed partial images and the intact sequences of an action is a key point of action prediction.

\subsubsection{Pose Generation}
\label{}
FAAST \cite{suma2011faast} provides a toolkit to create animated virtual characters using natural interaction from OpenNI-compliant depth sensors. Mesh body is also produced on basis of rigid limb motions and skinning weights both for humans and animals \cite{de2008automatic, chen2014adaptive}.

\textbf{Challenges:} Relations both inside skeletons of a pose and among the series of images need be deeply observed to generate precise poses.

\subsubsection{Pose Stripping}
\label{}

Radio frequency (RF) reflections of Wifi back from environment and humans is captured for estimation poses \cite{zhao2018through}. Heatmaps in both vertical and horizontal directions are parsed with encoders and then fused with keypoint confidence maps from RGB sequences, in which human skeletons can be stripped from background.

\textbf{Challenges:} The basic assumption is that the reflections from human body and other items are disparate. This assumption is susceptible to the things with the same reflection with human body.

\section{Datasets}
We conclude top 11 datasets which have high-frequent usage for skeleton approaches.

\textbf{1.NTU RGB+D Dataset \cite{shahroudy2016ntu}.}
This dataset consists of 56,880 action samples containing 4 different modalities of data for each sample: 1) RGB videos 136GB, 2) depth map sequences(including Masked depth maps 83GB and Full depth maps  886GB), 3) 3D skeletal data 5.8GB, 4) Infrared videos 221 GB, Total 1.3TB. In this dataset, the resolution of RGB videos are 1920 by 1080, depth maps and IR videos are all in 512 by 424, and 3D skeletal data contains the three dimensional locations of 25 major body joints at each frame.

\textbf{2.UT-Kinect Dataset \cite{xia2015robot}.}
This dataset includes two separate datasets. The first dataset(3.42G) is collected using Kinect mounted on top of a humanoid robot. There are 9 action types in the humanoid robot dataset:stand up,wave,hug,point,punch,reach,throw,run,shake hands.The second dataset(3.33G) is collected using a non-humanoid robot.There are 9 action types in the non-humanoid robot dataset:ignore, pass by the robot, point at the robot, reach an object, run away, stand up, stop the robot, throw at the robot, and wave to the robot. Each dataset contains 5 parts: 1) RGB images(.jpg), the resolution is 480x640. 2) Depth images(.png) the resolution is 320x240. 3) Calibrated depth images(.png), the resolution is 320x240. 4) Sketetal joint locations (.txt). Each row contains the data of one frame, the format is: frame number, frame count, skeletonId, (x,y,z) locations of joint 1-20. 5) Labels of action sequence (.txt).

\textbf{3.Florence 3D Actions Dataset \cite{seidenari2013recognizing}.}
The dataset collected at the University of Florence during 2012,has been captured using a Kinect camera. It includes 9 activities: wave, drink from a bottle, answer phone, clap, tight lace, sit down, stand up, read watch, bow.During acquisition,10 subjects were asked to perform the above actions for 2 or 3 times,which resulted in a total of 215 activity samples.

\textbf{4. Kinetics dataset \cite{ghanem2018activitynet}.}
This dataset stems from the competition of ActivityNet Large Scale Activity Recognition Challenge 2018, which started from 2016 CVPR. The dataset provided by Deepmind Team of Google currently includes a total of 600 categories and 500 thousand video clips all from Youtube. In the collected 600 categories, each one has at least 600 videos. Each video lasts about 10 seconds. The categories are classified into three main types: 1) Interaction between humans and objects such as playing musical instruments. 2) Human interaction such as handshake, hug. 3) Sports, etc. These three main types can also be described as Person, Person-Person, Person-Object.

\textbf{5. N-UCLA Dataset \cite{li2019skeleton, ghanem2010maximum}.}
Northwestern-UCLA dataset(N-UCLA)  was collected by three Kinect cameras, which contains 1494 sequences covering 10 action classes from 10 performers. And these 10 actions are: pick up with one hand, pick up with two hands, drop trash, walk around, sit down, stand up, donning, doffing, throw and carry. The subjects perform an action only one time in an action sequence, which contains an average of 39 frames. 

\textbf{6.SBU Interaction Dataset \cite{liu2017skeleton}.}
SBU Interaction dataset was collected with Kinect. It contains 8 classes of two-person interactions, and includes 282 skeleton sequences with 6822 frames. Each body skeleton consists of 15 joints.

\textbf{7.SYSU Dataset \cite{hu2019joint, hu2015jointly}.}
SYSU 3D Human-Object Interaction (SYSU) dataset is collected by Kinect camera. It contains 480 skeleton clips of 12 action categories performed by 40 subjects and each clip has 20 joints.

\textbf{8.MSR Action3D Dataset (publiced by Microsoft Research Redmond).}
MSR-Action3D dataset is an action dataset of depth sequences captured by a depth camera. This dataset contains twenty actions: high arm wave, horizontal arm wave, hammer, hand catch, forward punch, high throw, draw x, draw tick, draw circle, hand clap, two hand wave, side-boxing, bend, forward kick, side kick, jogging, tennis swing, tennis serve, golf swing, pick up $\&$ throw. It is created by Wanqing Li during his time at Microsoft Research Redmond.

\textbf{9.Berkeley MHAD Dataset \cite{chaudhry2013bio}.}
The Berkeley Multimodal Human Action Database (MHAD) contains 11 actions performed by 7 male and 5 female subjects in the range 23-30 years of age except for one elderly subject. All the subjects performed 5 repetitions of each action, yielding about 660 action sequences which correspond to about 82 minutes of total recording time. 

\textbf{10.UTD-MHAD Dataset \cite{chen2015utd}.}
This dataset was collected as part of our research on human action recognition using fusion of depth and inertial sensor data. For our multimodal human action dataset reported here, only one Kinect camera and one wearable inertial sensor were used. This was intentional due to the practicality or relatively non-intrusiveness aspect of using these two differing modality sensors. Both of these sensors are low cost, easy to operate, and do not require much computational power for the real-time manipulation of data generated by them. A picture of the Kinect camera can capture a color image with a resolution of 640 by 480 pixels and a 16-bit depth image with a resolution of 320 by 240 pixels. The frame rate is approximately 30 frames per second.

\textbf{11.HDM05 Dataset \cite{muller2007documentation}.}
HDM05 dataset is a motion capture database which contains more than three hours of systematically recorded and well-documented motion capture data in the C3D as well as in the ASF/AMC data format. Furthermore, HDM05 contains for more than 70 motion classes in 10 to 50 realizations executed by various actors. The HDM05 database has been designed and set up under the direction of Meinard Müller Tido Röder, Michael Clausen, Bernhard Eberhardt, Björn Krüger, and Andreas Weber. The motion capturing has been conducted in the year 2005 at the Hochschule der Medien (HDM), Stuttgart, Germany, supervised by Bernhard Eberhardt.

\section{Conclusion}

Skeleton-based approach as a significant part was evolving along with the blooming development of artificial intelligent applications (such as object detection, action identification, pose estimation, and so on) which had attracted high attentions. This paper observed skeleton-based approaches and categorized these techniques in accordance with target tasks rather than theoretical frameworks, which is useful for introducing this scope.


%




\ifCLASSOPTIONcaptionsoff
  \newpage
\fi




\bibliographystyle{IEEEtran}
\bibliography{IEEEabrv,bare_jrnl} 

\end{document}